# Defining Digital Quadruplets in the Cyber-Physical-Social Space for Parallel Driving

Teng Liu, Yang Xing, Long Chen, Dongpu Cao, Fei-Yue Wang

*Abstract*—Parallel driving is a novel framework to synthesize the vehicle intelligence and transport automation. This article aims to define the digital quadruplets in parallel driving. In the cyber-physical-social systems (CPSS), based on the ACP method, the names of the digital quadruplets are first given, which are descriptive, predictive, prescriptive and real vehicles. The objectives of the three virtual digital vehicles are interacting, guiding, simulating and improving with the real vehicles. Then, the three virtual components of the digital quadruplets are introduced in detail and their applications are also illustrated. Finally, the real vehicles in parallel driving system and the research process of the digital quadruplets are depicted. The presented digital quadruplets in parallel driving are expected to make the future connected automated driving safety, efficiently and synergistically.

*Index Terms*—Digital quadruplets, Parallel driving, Descriptive vehicles, Predictive vehicles, Prescriptive vehicles, Real vehicles

## I. Introduction

Artificial intelligence technologies enable the fleet development of the autonomous vehicles. Many researches have discussed how to build an efficient and safe self-driving car. In general, four function modules are included in an autonomous vehicle, which are prescription, decision-making, planning and control. From DARPA challenge to current vehicle manufacturers, many schemes have proposed to construct a self-driving car with high automation.

There is still a long way to go in the development of L5 automated driving. Human drivers are still the main participants in the common transportation. Hence, the autonomous vehicle should consider the driving intentions of the human drivers, and then conduct an appropriate driving task. To enhance the ability of automated cars, a combination of virtual and real driving framework is proposed and shown in the 29th IEEE International intelligent vehicle conference (IEEE IV 2018). It is called parallel driving.

Motivated by the ACP method, there are two "vehicles" (virtual and real vehicles) exist in parallel driving. The virtual vehicles are the ideal modeling of the real vehicle. It could record all the driving data of the real vehicle and guide the operation of the real vehicle. The interaction between the virtual and real vehicle is the biggest characteristic of the parallel driving. With the help of virtual vehicle, the real vehicle is able to improve efficiency and safety, and also promote the whole system.

In the first day's plenary lecture of IV 2018, one keynote speaker gave a fantastic speech on parallel driving, wherein the framework, theoretical development and applications of parallel driving were introduced in detail. Parallel driving is a cloud-based cyber-physical-social system (CPSS) framework, which was presented in 2004. It had been developed since then to enhance the performance and safety of connected automated vehicles. Many advanced theoretical technologies have been proposed in parallel driving, such as parallel vision/perception, parallel reinforcement learning and parallel planning/control.

In the last day of IV 2018, the International Parallel Driving Alliance (iPDA) orgnized the International Intelligent Vehicle Joint Road Demonstration. The purport is "From Parallel Driving to Smart Mobility". Fours parts are displayed in the road demonstration, which are the real-time status detection by the central driver, responsive takeover in general traffic scenario, active obstacle avoidance and the the active takeover in emergency traffic scenario.

In this article, we present a novel concept for parallel driving, which is called digital quadruplets. This system consists of the individual or connected real vehicles and three virtual 'guardian angels' for them. The three virtual vehicles are developed in the artificial world and they are named as descriptive vehicle, predictive vehicle, and prescriptive vehicle. The descriptive vehicle focuses on modeling the real vehicles accurately. The predictive vehicle aims to address the perception, decision-making, planning and control for the connected automated vehicles using different computation experiment techniques. Furthermore, the objective of the prescriptive vehicle is handling the communication of real vehicle and virtual vehicles.

To introduce the digital quadruplet of parallel driving clearly, the remaining paper is arranged as follows. Section II describes the framework of parallel driving and the structure of digital quadruplets in parallel driving. Three virtual 'guardian angels' in the artificial world of parallel driving are defined elaborately in Section III, IV and V, respectively. In Section VI, the development of real vehicles in the parallel driving system and the research process of digital quadruplets are discussed.

T. Liu, Y. Xing, and L. Chen are with Vehicle Intelligence Pioneers Inc., Qingdao Shandong 266109, China, and also with Qingdao Academy of Intelligent Industries, Qingdao Shandong 266109, China. (email: tengliu17@gmail.com, y.xing@cranfield.ac.uk, chenl46@mail.sysu. edu.cn).

D. Cao is with the Department of Mechanical and Mechatronics Engineering, University of Waterloo, N2L 3G1, Canada. (email: dongpu@uwaterloo.ca)

Fei-Yue Wang is State Key Laboratory of Management and Control for Complex System, Institute of Automation, Beijing, China (e-mail: feiyue@ieee.org).

## II. DIGITAL QUADRUPLETS IN PARALLEL DRIVING

This section tends to define the digital quadruplets in cyber-physical-social space (CPSS) for parallel driving. First, the CPSS and ACP method are introduced in detail. Based on these two theories, the framework of parallel driving is given. Finally, the digital quadruplets in the parallel driving are discussed.

### A. ACP Method and CPSS

The ACP method means the artificial societies (A), computational experiments (C) and parallel execution (P), which was presented by Fei-Yue Wang since 2004 [1-4]. Especially, A is usually used for modeling the complex systems, C is applied to calculate and analyze the responses and P is utilized for control and management. These three components of ACP can be mapped into three parallel worlds that are called physical, mental and artificial worlds, as shown in Fig. 1. In a parallel system, ACP can be activated both in the artificial system and real system to handle the complex control problems [5, 6].

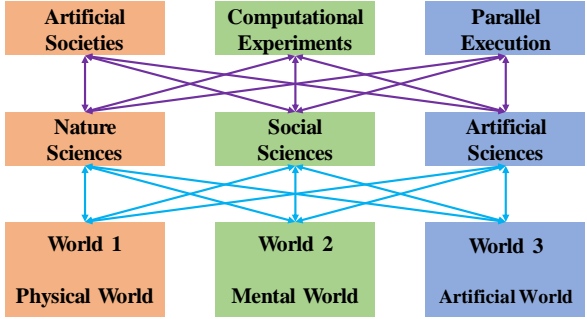

Fig. 1. Mapping ACP method into three parallel worlds.

With the development of the artificial intelligence, big data and communication technologies, the CPSS attracts more and more attention in recent years [7]. By considering the human and social characteristics in the cyber-physical systems (CPS), the complex systems can be designed and operated more efficiently and reasonably. The critical elements in the CPSS also contain the physical, mental and artificial worlds. It indicates that the ACP method can optimize and manage the complex system problems in CPSS via exploiting knowledge in the artificial world to guide the mental and physical world [8].

### B. Parallel Driving

Due to the convenience, safety and efficiency, autonomous vehicles are experiencing an extreme development in scientific and industrial fields [9, 10]. However, the individual intelligent vehicle usually fails to recognize the social nature on the road. It is difficulty for this vehicle to understand the meaning of other vehicles' behaviors and to figure out how to communicate with human drivers. Hence, CPSS-based parallel driving is presented to achieve the best trade-off between reliability and cost [11], see Fig. 2 as an illustration.

Different from the ordinary automated driving, parallel driving includes the artificial driver and artificial vehicle. The parameters and information of the real and artificial vehicles co-exist into the three parallel worlds [12]. Specifically, the real driving exists in the physical world, which consists of the physical behavior of the real vehicle and real driver. The cognitive behaviors of the real and artificial drivers locate in the mental world, including driver attention, intention and attributes. There are two layers in the artificial world, the first layer is the artificial driver and artificial vehicle, and the second layer is the information of people, location and technologies [13].

The modeling of the vehicle, driver, environment and so on will be built in the artificial world. The artificial vehicles can be simulated in the uncommon and emergency scenarios, and the generated commands could guide the real vehicles in physical world. Furthermore, the data from the real vehicle can be applied to improve the accuracy of the artificial modeling. These processes can be realized by the special computational experiment and parallel execution methods [14]. Different types of the virtual vehicles can co-exist in the artificial world to enhance each other and communicate with the real vehicle, which are discussed in the next section.

### C. Digital Quadruplets in Parallel Driving

In parallel driving, we define three virtual vehicles, which are named descriptive, predictive and prescriptive vehicles. These vehicles can be regarded as the 'guardian angels' of the real vehicles in parallel driving. Combining with the real autonomous vehicle in parallel driving, they are called the digital quadruplets of parallel driving. As depicted in Fig. 3, different real vehicles in the physical world can suggest various automation levels, from Level 0 to Level 5. They interact with each other by communication technologies to achieve safe and efficient management [15] (e.g., fleet management [16]).

Based on the historical data and current observations, the first guardian angel, descriptive vehicle aims to describe the operation of real vehicles [17] accurately. For example, how to model the autonomous vehicle powertrain, how to establish multiple scenarios for real vehicles, how to build the environment modeling and what data should be selected for communication between physical and artificial world. By settling these problems, this virtual vehicle can form as a self-consistent system to represent the real driving and real vehicle in the artificial world.

The second virtual guardian angel, the predictive vehicle can be treated as multi-agent learning systems, which use diverse computation experiment approaches (e.g., machine learning [18] and software analytics [19]) to realize self-calibrating and self-sublimating. The first step is self-calibrating, based on the modeling, data and knowledge from the descriptive vehicles and real vehicles, the predictive vehicles can improve the generative virtual models and evaluate the real physical models. For example, perform the same controls into the descriptive vehicles and real vehicles and make them output the similar states. The second step is self-sublimating, using the artificial intelligence (AI) technologies to make all the virtual vehicles traverses different testing scenarios, and then boost themselves and guide the real vehicles. For example, the emergency scenarios, the limiting conditions and the polytropic weather.

Finally, the third guardian angel, prescriptive vehicle decides two aspects of problems, the first one is whether the controls learned from the predictive vehicles can be used into the physical world, and the second one is how the virtual vehicles interact with the real ones. To address the first problem, the safety-aware, behavior-aware and situation-aware-based learning methods (e.g., apprenticeship learning [20]) can be leveraged to evaluate the controls, policies and strategies from predictive vehicles. To solve the second problem, the human-machine interface techniques, such as fingerprint recognition, voice interaction and touch operation can be selected to link the different types of virtual and real vehicles. In the following Sections, we introduce the details of the four digital vehicles in parallel driving, as well as their corresponding applications.

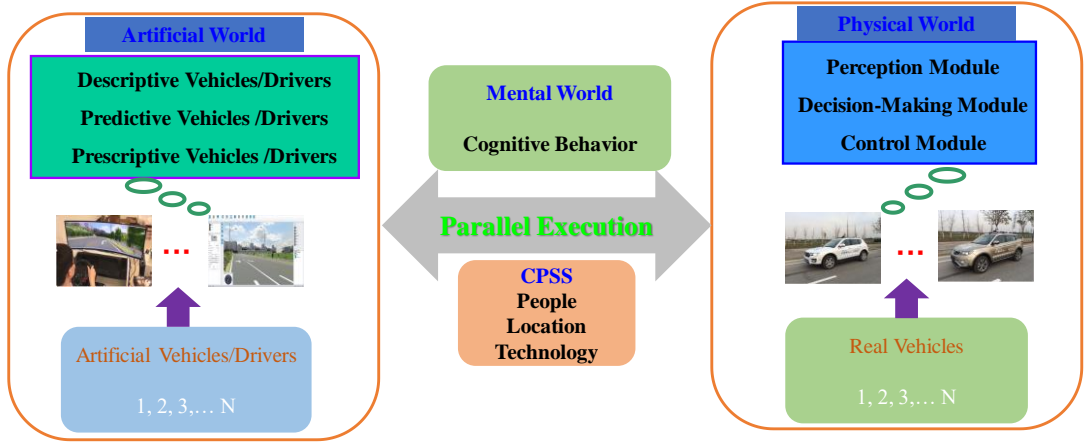

Fig. 2. Framework of the CPSS-based parallel driving.

## III. DESCRIPTIVE VEHICLE AND APPLICATIONS

### A. Design Descriptive Vehicle Based on Descriptive Learning

The first component of the digital quadruplets is called the descriptive vehicle. Its construction relies on the descriptive learning theory [21]. The nature behind the descriptive learning is to make strategies that can describe how the learning is generated and how to learn a model that can be used to mimic and describe the real-world vehicle. The descriptive vehicle is one of the core components in parallel driving which is responsible for learning of real-world vehicle dynamics based on limited prior knowledge and observations.

According to the study [22], the descriptive learning process is to design a self-consistent system that does not violate the principles and rules behind the naturalistic object. Specifically, the descriptive vehicle has to learn and form itself according to its observation on the real-world system. The descriptive vehicle can be represented as:

$$\begin{cases} s_i = F(a_i), i = 0,1... \\ a_j = P(s_i), i = 0,1... \end{cases} \quad (1)$$

where $a_i \in A$ and $s_i \in S$ represent the action and state within the action set $A$ and state set $S$. $F(\cdot)$ is an inductive function that determines how the action taken by the agent influence the state, while $P(\cdot)$ describes the policy that has to make according to the state.

The $F(\cdot)$ and $P(\cdot)$ can be optimized according to the following principle:

$$\begin{cases} F(a_i) \triangleq \arg\max_{s_j \in S} L(a_i, s_j) \\ P(s_i) \triangleq \arg\max_{a_j \in A} R(s_i, a_j) \end{cases} \quad (2)$$

where $L_s(a_i, s_j)$ represents the likelihood that action $a_i$ and $s_i$ happened sequentially, and $R_a(s_i, a_j)$ measures the real long term reward according to the state and action. In sum, the function $F(\cdot)$ is determined by the probability that some observable states transfer according to the actions, and the policy function $P(\cdot)$ depends on the reward of the action series with respect to the state sequence.

As aforementioned, the descriptive vehicle is to form a self-consistent virtual system which does not break the law with respect to the observation. The construction process for the descriptive vehicle can be shown as:

$$\begin{aligned} \arg\min d_{F,P} &= (|s_i' - F_A(P(s_i'))|, |a_i' - P_A(F_A(a_i'))|) \\ i &= 0,1,...m \\ s.t.\ g_f(|s_j - F_A(a_j)|) &\leq 0 \\ g_P(|a_j - P_A(s_j)|) &\leq 0,\ j = 0,1,...,n \end{aligned} \quad (3)$$

where $d_{F,P}$ measures whether the mutual-dependency of the state and action hold consistent, $g_f$ and $g_P$ are the constraint upon the induction and policy function. By minimizing this cost function, a self-consistent descriptive vehicle can be constructed. More detail about the descriptive learning please refer to [17].

The descriptive vehicle in the artificial world needs to learn the pattern of the dynamics and behaviors of the real vehicle. The descriptive vehicle is parallel to the real vehicle and will run asynchronously. The descriptive vehicle is an active counterpart of the real vehicle, which provides a playground to explore the space of the state and potential response of the real agent in a more efficient and less harmful manner. The descriptive vehicle is built based on the descriptive learning theory. Inspired by the descriptive learning theory in [23], the following part will introduce three potential learning methods, namely, the behaviourist, cognitive, and constructivist learning in a more general fashion.

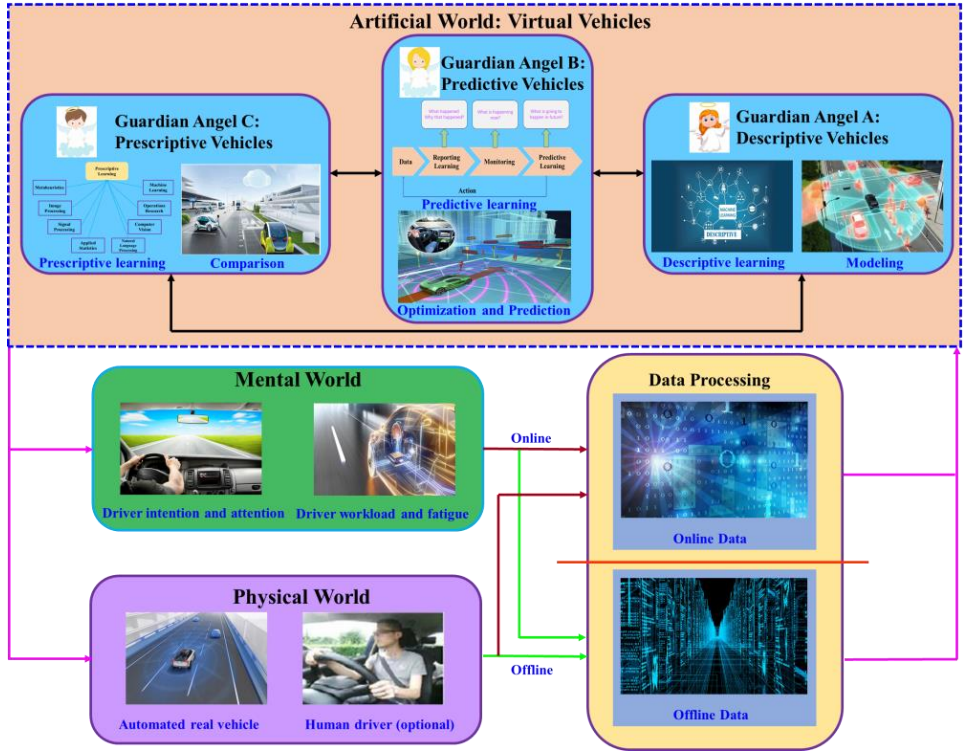

Fig. 3. Framework of the digital quadruplets in parallel driving.

*B. Behaviorism Learning Method*

Behaviorism learning explains the behaviour-based learning process, which trains the artificial agents with the observed stimuli-response pairs. The behaviourism learning does not refer the mental status of the real agent and only try to learn the optimal condition-action associations. As shown in Fig. 4, the behaviourism learning only focus on the physical world in the lowest level of parallel driving system. The behaviourism learning in the descriptive learning is a kind of behaviour cloning within the supervised learning scope. Many studies have tried to devise an artificial model based on this method [24].

The behaviorism learning scheme is the most common way to construct a simple descriptive vehicle. According to the measured input and output, the dataset will form the training date for the model training and supervised learning method can be directly applied to train the model. For example, in [25], the authors tried to model and estimate the braking pressure based on the naturalist data collected from the CAN bus, this braking assistance system can be viewed as a sub-system of the descriptive vehicle. In [24], the authors proposed an end-to-end learning framework to construct a holistic control unit for the autonomous vehicle. The inputs are images from multiple cameras and the output are the steering angle and the braking pressure. This is a typical behavior cloning process based on the observable stimulus-response associations.

*C. Cognitive Learning Method*

As shown in Fig. 4, the human driver exists in both the physical world as well as the cognitive world. For the fully autonomous driving vehicle, the driver related physical and cognitive behaviors are not necessary, however, if the automated vehicle has to disengage or the driver intent to take-over the control, the driver cognitive and physical behavior will re-engage in the driving loop. Therefore, to construct the holistic quadruplets driving system, the driver mental process has to be identified and involved.

The cognitive process of the human driver contains the attention, intention, workload and the residual driving capacity. The patterns behind the mental status of the human driver cannot be learned directly through the behavior cloning method as described in the last section. The primary object of cognitive psychology and cognitive learning is focused on the inference of inner mental states. The mental states inference depends on the long-term cognitive process and the corresponding outer context as well as the driver behaviors.

As the driver mental states cannot be measured directly, the only way to understand the human driver is to infer the states based on the traffic context and driver behaviours. Meanwhile, cognitive learning relies on the long-term dependency of the behaviour information rather than the instance signal. For example, in [26], the driver lane change and turn intention inference framework were designed based on the analysis of driver behaviours in a certain period. The long short-term memory (LSTM) based recurrent neural network (RNN) is used to store the driver behaviour sequence and make inference accordingly. In [27], an electroencephalogram (EEG) based driver drowsiness detection system was proposed by combining independent component analysis (ICA) and power-spectrum. A two-second sliding window of the EEG signals are chosen to infer the driver drowsiness states.

## IV. PREDICTIVE VEHICLE AND APPLICATIONS

This section discusses the first guardian angel, the predictive vehicle in parallel driving. First, the definition and the necessity of predictive learning is introduced. Then, the framework of the predictive learning-enabled virtual vehicles is demonstrated. Finally, one example of the prediction and planning processes in PVs are depicted.

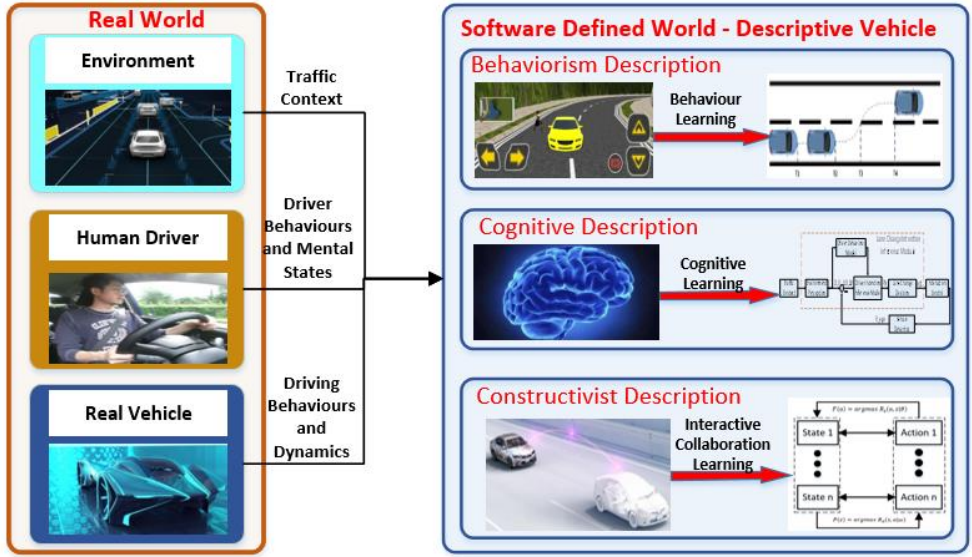

Fig. 4. Descriptive vehicles in the artificial world of parallel driving.

*A. Predictive Learning*

To be an intelligent vehicle of high automation levels, the onboard controller need to perceive the states of the world (environment and itself) first so as to make accurate predictions and planning [28]. Then, the controller is able to update the estimates of these states and record them. Finally, the computation experiment methods are applied to predict what actions will lead to the desired states of the world. By completing these steps expertly and precisely, the vehicle becomes more and more intelligent.

In parallel driving, predictive learning means predicting any part of the historical, current and future information to enrich the modeling or algorithms. For example, predicting the consequences of a sequence of control commands provided by the onboard controller. This definition of predictive learning is also mean by unsupervised learning [22] and has been researched for several years. The relevant approaches are adversarial learning [29], adjoint state [30], predictability minimization [31], actor-critic reinforcement learning [32] and Dyna architecture [33].

*B. Predictive Vehicles*

Fig. 5 shows the framework of the predictive vehicles in the artificial world. In the upper level, the real vehicles take observations from the environment and feedback the actions. The states and actions from the real vehicles affect the defined objective, which is expected to be minimized during long-term interval.

To predict and plan ahead using predictive learning, the real vehicles are simulated as many predictive vehicles, as shown in the lower level. Three parts exist in each predictive vehicle, and the computation module also acquires inferred observations from the simulated environment. The virtual state and action come into the experiment module and outputs the predicted cost. Different predictive vehicles result in various objective cost, which is regarded as knowledge. These knowledges can be used to direct the real vehicles and also improve the generative modeling in descriptive vehicles.

As stated above, the predictive vehicles can achieve different purposes in prediction and planning perspectives. In the following sub-sections, one example of predictive vehicles is introduced: the fuzzy encoding module.

*C. Fuzzy Encoding Module*

Numerous states in the vehicles change with the time, and thus they can be recorded as a vector. To predict each future state based on the historical data, a finite Markov chain (MC) [34] is used to model this variable as $\{x_i \mid i=1, \ldots, N\} \subset X$, where $X \subset R$ is bounded. Maximum likelihood estimator (MLE) is applied to calculate its transition probability as follows [35]

$$\begin{cases} p_{ij} = P(x^+ = x_j \mid x = x_i) = \dfrac{M_{ij}}{M_i} \\ M_i = \sum_{j=1}^{N} M_{ij} \end{cases} \quad (4)$$

where $x$ and $x^+$ are the current and one-step ahead states, $p_{ij}$ is the transition probability from $x_i$ to $x_j$. $M_i$ is the total transition times from $x_i$ and $M_{ij}$ is the counts from $x_i$ to $x_j$.

For special discrete states, all the transition probability $p_{ij}$ can constitute a matrix, $\Pi$. Then, $n$-steps ahead states can be expressed as vector form

$$(p^{+n})^T = p^T \Pi^n \quad (5)$$

In the fuzzy encoding module (FEM), the set $X$ is dispersed as several fuzzy subsets $\psi_i$, $i=1, \ldots, N$. This subset denotes a pair $(X, \theta_i(\cdot))$, wherein $\theta_i(\cdot)$ is a Lebesgue measurable membership function that means

$$\theta_i : X \to [0,1] \, s.t. \, \forall x \in X, \exists i, 1 \leq i \leq N, \, \theta_i(x) > 0 \quad (6)$$

where $\theta_i(x)$ describes the degree of membership of $x \in X$ in $\theta_i$. It can be discerned that a continuous state $x \in X$ in fuzzy encoding may refer to several states $x_i$ in MC model [36].

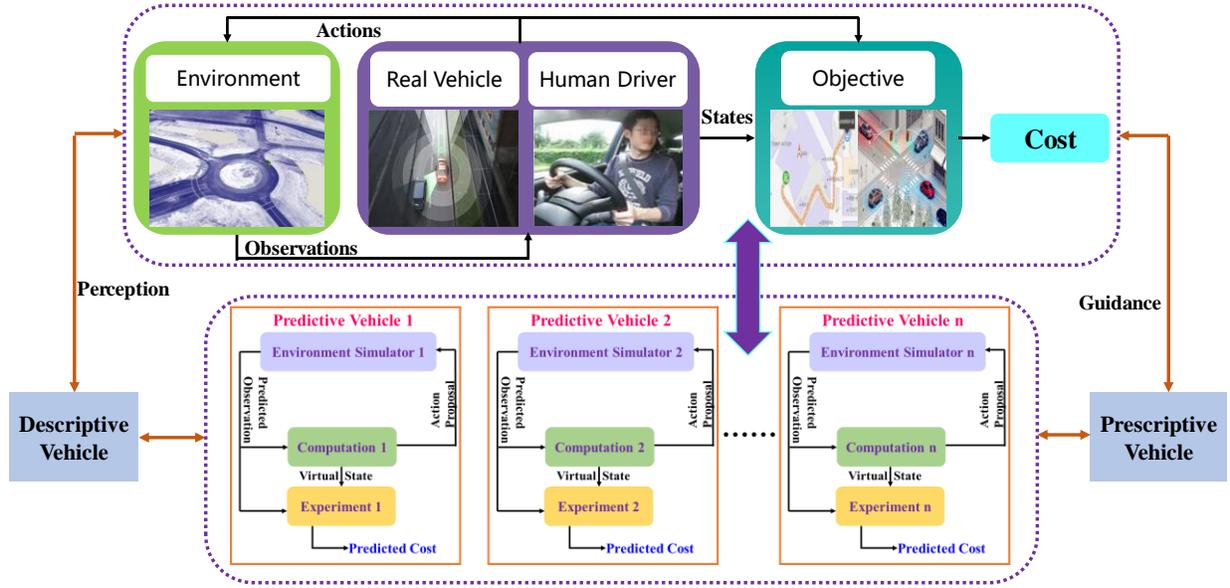

Fig. 5. Predictive vehicles in the artificial world of parallel driving.

To predict future states in FEM, two transformations need to be constructed via approximate reasoning [37]. For each state $x \in X$, a $N$-dimensional possibility vector is mapped to them in the first transformation

$$K^T(x) = \theta^T(x) = [\theta_1(x), \theta_2(x), \cdots, \theta_N(x)] \quad (7)$$

where the $K(x)$ is a possibility vector, so the sum of its elements may not equal to 1. This transformation maps the state $x \in X$ to a vector in possibility space $X_1$ and it is named fuzzification. Then, the second transformation maps the possibility vector $K(x)$ as a probability vector $K_1(x)$ via normalization

$$K_1(x) = K(x) / \sum_{i=1}^{N} K_i(x) \quad (8)$$

Through the second, a possibility space $X_1$ is connected to a probability space $X_2$. Then, the next state in $X_2$ is depicted as follows

$$(K_1^+(x))^T = (K_1(x))^T \Pi \quad (9)$$

Hence, the transition probability also indicates a transition from $\psi_i$ to $\psi_j$. Then, the one-step ahead state in $X$ can be encoded in $X_2$ as

$$\omega^+(x) = (K_1^+(x))^T \theta(x) = (K_1(x))^T \Pi \theta(x) \quad (10)$$

Thus, the one-step ahead vector can be represented by the expected value over the possibility vector as

$$\begin{cases} x^+ = \int_X \omega^+(y) y \, dy / \int_X \omega^+(y) \, dy \\ \int_X \omega^+(y) y \, dy = \sum_{j=1}^{N} K_{1,j}(x) \sum_{i=1}^{N} p_{ij} \int_X y \theta_i(y) \, dy \\ \int_X \omega^+(y) \, dy = \sum_{j=1}^{N} K_{1,j}(x) \sum_{i=1}^{N} p_{ij} \int_X \theta_i(y) \, dy \end{cases} \quad (11)$$

The volume and centroid of the membership function $\theta_i(x)$ is expressed as

$$\begin{cases} \overline{c}_i = \int_X y \theta_i(y) \, dy \\ V_i = \int_X \theta_i(y) \, dy. \end{cases} \quad (12)$$

Combining Eq. (11) and (12), the one-step ahead state of $x$ in $X$ is denoted as follows

$$\begin{cases} x^+ = \dfrac{\sum_{j=1}^{N} K_{1,j}(x) \sum_{i=1}^{N} p_{ij} V_i \overline{c}_i}{\sum_{j=1}^{N} K_{1,j}(x) \sum_{i=1}^{N} p_{ij} V_i} \end{cases} \quad (13)$$

The above calculation process represents one virtual predictive vehicle, and the computation experiment method is denoted by FEM. This virtual vehicle can predict many states of the real vehicles based on the historical data, such as vehicle speed, power demand, engine speed, motor current and so on. By doing so, the vehicles can adjust the control actions according to the prediction and thus run more safely and efficiently [38].

## V. PRESCRIPTIVE VEHICLE AND APPLICATIONS

### A. Prescriptive Vehicle Concept

The prescriptive vehicle is responsible for guiding the real vehicle to make decisions and achieve the target outcomes, which aims to generate a series of proper action strategies to help the real vehicle deal with the complex traffic situation. The prescriptive vehicle is concerned about whether the policy learning in the parallel system can be properly utilized in the real world.

By and large, the prescriptive vehicle can be treated as a special kind of predictive vehicle. However, the prescriptive vehicle must be smarter than the predictive vehicle since the prescriptive vehicle has to learn the long (or short) term dependency between the taken actions and the response in case to guide the real vehicle when facing complex situations. As the predictive vehicle enables the predictions of what will happen in the future, the prescriptive vehicle can take the predictions from multiple parallel predictive vehicles to evaluate which action contributes to the maximum rewards and generate the target consequence [39].

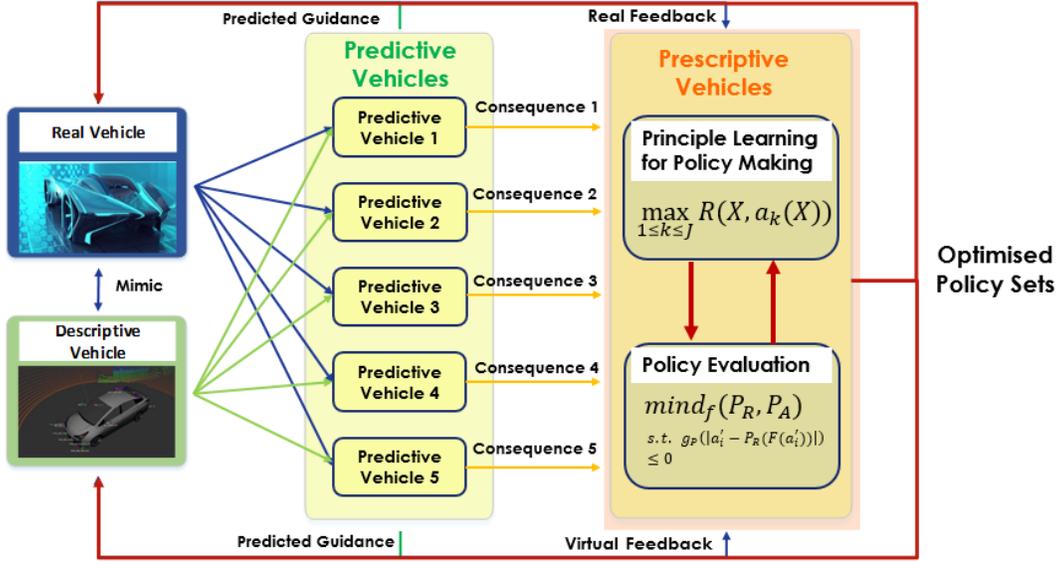

Fig. 6. Predictive vehicles in the artificial world of parallel driving.

According to the aforementioned parallel quadruplet framework, the prescriptive vehicle has two important tasks. The first one is to judge whether the learned and predicted action can be adopted by the real vehicle. Secondly, how to efficiently interact with the real vehicle. In the following, these two problems will be addressed based on the parallel driving theory.

*B. Learning Process of the Prescriptive Vehicle*

After the predictive vehicle obtained the knowledge of how the states of the real vehicle and virtual vehicle transformed after taken the actions, the prescriptive vehicle will go on a step further to concern about whether the policy learned in the parallel system can be applied to the real vehicle or not. Based on the concept of the prescriptive vehicle, we divided the construction of the prescriptive vehicle as a two-stage learning process. The first stage is to learn the principle of taking actions, while the second stage is to evaluate the candidate actions which are likely to happen and find the optimised policy sets to suggest the real vehicle, the general framework of prescriptive vehicle and its relationship with the other components of the digital quadruplet can be found in Fig 6.

As shown in Fig. 6, the prescriptive vehicle will take the predicted consequences from the predictive vehicle as inputs, then it will try to learn the principle of making policy based on these consequences with necessary context information from both the real world and the descriptive vehicle. After the prescriptive vehicle has obtained the optimal strategies for guiding the real vehicle, it will send this information to the real vehicle and the descriptive vehicle to guide the decision making.

Meanwhile, the prescriptive vehicle will feedback the learned principle and policy to the predictive vehicle to guide its learning direction in case to make more reasonable prediction. According to this, the first stage in the prescriptive vehicle is to learn how to make reasonable policy according to the predicted consequence and the environment, which can be described as [17]:

$$\max_{a_k, 1 \leq k \leq J} R(X, a_k(X)) \quad (14)$$

where $X=\{x_i\}$, $i=1,\ldots,I$ is the data set, and $A=\{a_k(X)\}$, $k=1,\ldots,J$ is the action set that we can execute. The first learning stage is to maximise the reward function $R(\cdot)$ so that the prescriptive vehicle can learn the principle of making decision.

In the second stage, based on the learned principle, the prescriptive vehicle will update its parameter to make sure with a given state $s_i \in S$, the action $a_i \in A$ taken by the real vehicle should be close to the action given by the prescriptive vehicle. Hence, the second stage can be represented as an optimization problem with the following cost function:

$$\min d_F(P_R, P_A)$$
$$s.t. \ g_P(|a_i' - P_R(F_A(a_i'))|) \leq 0, i=1,\ldots,m \quad (15)$$

where the objective function $d_f$ measures the difference between $P_R$ and $P_A$ conditional to the generative model $F$.

Nowadays, the reinforcement learning enables us to solve the first stage problem in a very efficient manner. However, many works are still expected to be done in the second stage, which relies on generating new data and use these data to explore the untouched space in both real world and parallel world.

## VI. RESEARCH PROCESS OF DIGITAL QUADRUPLETS

In this section, we introduce the research process of the digital quadruplets in parallel driving. First, the real vehicles are explained, including the sensors, the function and the relevant algorithms. Then, the study about the first guardian angel, descriptive vehicle, is depicted. The research focuses on the identification and analysis of the driver posture or workload in autonomous vehicles. Finally, an example of the predictive vehicle is given to display an epitome on the research of the second guardian angel.

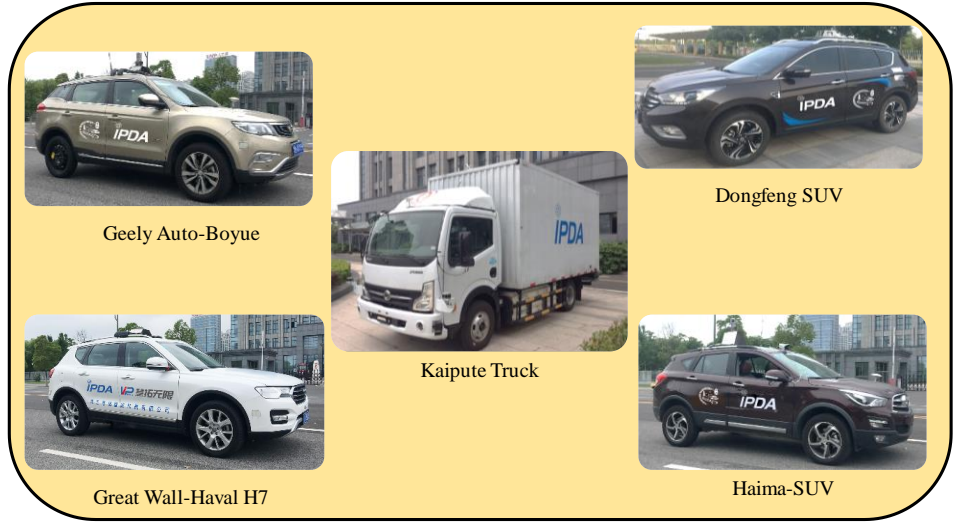

Fig. 7. Different real vehicles in parallel driving system.

## A. Real Vehicles

The real vehicles in parallel driving can be different automation levels of automated vehicles. The number of them is variable. This vehicle can be an individual one in the parallel driving system and interacts with the virtual vehicle in the artificial world. Furthermore, they also can constitute a vehicle fleet and communicate with each other through vehicle-to-vehicle (V2V) technologies.

Fig. 7 shows different kinds of real vehicles in the existing parallel driving system. Many sensors and equipment are installed in each vehicle, including laser radar, human-machine interface (HMI), two cameras, industrial control computer (ICP), wireless communication and emergency stop (E-stop). Parallel to each vehicle, a virtual vehicle is modeled in the artificial world and it can achieve all the functions as the real vehicle does.

Three main functions are integrated in the ICP of real vehicle, which are planning and decision-making, perception and control. Based on the data from sensors, the perception module can construct the mapping of the road and environment and recognize the pedestrian and obstacles, then the decision-making module could decide the special behaviors, such as accelerating, changing lane and stopping, finally the control module may execute these commands via controlling the components of the vehicle. Furthermore, the vehicle can be controlled by the tele-control system, which is determined by the cloud management system. Also, the commands from the prescriptive vehicle in the artificial world can be applied to guide the real vehicle.

## B. Driver Modeling in Descriptive Vehicle

Based on the descriptive vehicle, the construction of the descriptive vehicle/driver can be realized based on three methods, which are behaviorism learning, cognitive learning, and constructivist learning process. In this part, the application of behaviorism learning, and cognitive learning will be introduced with the demonstration of driver behavior identification and driver mental intention inference, respectively.

The behavioral modelling for descriptive vehicle and descriptive driver is simpler than the cognitive modeling and constructivist modeling as it only requires capturing and modeling the observable outer behaviors.

Fig. 8. A deep learning-based driver behavior recognition.

Fig. 8 indicates the driver behavior identification based on a deep learning approach [40]. The end-to-end driver behavior modeling procedure can be described as follows. Firstly, the color images are cropped and segmented with a Gaussian mixture model (GMM). Then, the segmented images are fed into a pretrain deep convolutional neural network (CNN), which is trained with the transfer learning scheme and the driver behavior dataset. Finally, the CNN model can predict the driver behaviors.

The driver behaviors include some common in-vehicle activities such as normal driving, mirror checking, phone answering, texting, and use radio/GPS devices. Once the driver behavior recognition model is trained, the driver behavior pattern can be understood. Then, the virtue drivers which acts like the real human drivers to generate virtual behavior sequences can be built with the assistance of deep learning model such as the generative adversarial network (GAN). Similar approach can be applied to the construction of the behavior learning-based vehicle model, which focus on the learning of outer vehicle dynamics and driving behaviors.

Next, the cognitive learning process with respect to driver intention inference is described. Unlike the behavior modelling process, the cognitive modeling process for the human driver requires capturing the long-term dependency between the be-havior and traffic context so that the mental status can be inferred. Therefore, the cognitive modeling is more complex than the behavior modeling. The driver behavior modeling can be regarded as a sub-component of the driver cognitive modeling process. Taking the driver intention inference system for example, the intention cannot be measured directly, instead, it only can be inferred according to the driver outer behaviors, traffic context, and the continuous control to the vehicles [41].

The cognitive process for driver intention can be summaries as: firstly, the traffic context will stimuli the human driver to generate intentions. Then, to realize the intention, human driver must take a series of context perception and control actions. Once the driver decides to finish the intention, a series of vehicle control actions will be taken. Therefore, the modeling of driver mental status has to rely on a holistic model which takes the traffic context, driver behaviors, and vehicle dynamic information into consideration.

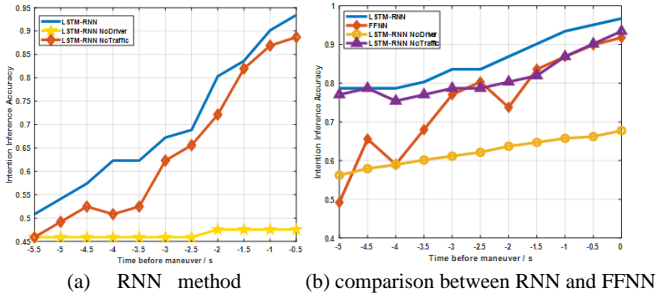
(a) RNN method (b) comparison between RNN and FFNN
Fig. 9. Driver intention using recurrent neural network and comparison.

Fig. 9 illustrates the prediction of driver lane change intention based recurrent neural network with respect to different prediction time. Fig. 9 (a) indicates the lane change intention inference results given by RNN with different input signals. The RNN that consider both the traffic context and driver behavior information (blue line) give the best prediction results and the lane change intention prediction achieved 96% accuracy when estimate the intention 0.5 seconds prior the maneuver happens. Fig. 9 (b) shows the comparison between RNN and feedforward neural network (FFNN) with respect to different prediction horizon. The RNN gives much better prediction accuracy compared with the FFNN as it captures the long-term dependency between the driver behaviors and traffic context.

By using machine learning models, driver intention can be predicted precisely even before the maneuvers being initiated [41]. Like the construction of virtual behavioral learning-based driver, the virtual cognitive learning-based driver can be realized with GAN model and trained with both real-world and virtual-world driver and traffic context data. This will enable the descriptive driver has the ability to modeling and mimic both the behavioral and cognitive dynamics of the real driver.

*C. Parameter Prediction in Predictive Vehicle*

Based on the modeling in descriptive vehicle, the algorithms in predictive vehicle could forecast different parameters for the autonomous vehicles. For each individual vehicle, we describe the power demand prediction based on the FEM provided in Section IV. C. To regard several real vehicles as vehicle fleet, we explain how to use model predictive control (MPC) to decide longitudinal speed of each vehicle to reduce red light idling.

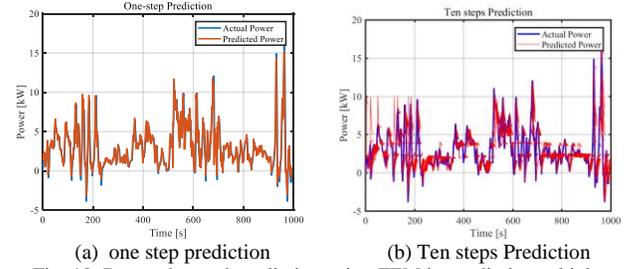
(a) one step prediction (b) Ten steps Prediction
Fig. 10. Power demand prediction using FEM in predictive vehicle.

Power demand changes according to the variation of traffic conditions, driver intention and road grade. And it decides the energy efficiency and control rules of the onboard energy sources. Hence, it is necessary to predict the power demand for autonomous vehicles when design the adaptive control algorithms. In the predictive vehicle of parallel driving, power demand can be known in advance using FEM. This information can be reference for real vehicle control.

Fig. 10 depicts the FEM-based predicted power demand. For different applications, the steps of future power demand are different. In Fig. 10 (a), we show the one-step ahead power demand. The information can be used for energy management, longitudinal control and planning. The long-predicted data (10 seconds) is described in Fig. 10 (b), which is beneficial for decision-making and fleet management. In additional, the proposed FEM can be applied for many other parameters prediction, such as acceleration, braking force, torque and rotational speed. By doing this, the predictive can enhance the virtual modeling in descriptive vehicle and guide the operation of real vehicles [42].

Furthermore, reducing stopping at red lights could improve the energy efficiency of automated vehicles. In the framework of predictive vehicle, the velocity of a vehicle fleet can be determined via MPC algorithm. Assuming the information of speed, signal lights and distances can be obtained by communication technologies. The computed vehicle speed is displayed in Fig. 11. Fig.11(a) provides by benchmark method, wherein the speed of each vehicle is always equal to zero. This means the relevant vehicle need to stop to wait for the red light. However, in Fig.11(b), the proposed method can make all the vehicles pass the signal light smoothly. This policy can improve the energy efficiency for autonomous vehicles and prevent from collision.

In the predictive vehicle, much more algorithms and approaches can be used to improve safety and efficiency for the autonomous vehicles. And the communication process between the virtual vehicle and real vehicle is mainly conducted by the prescriptive vehicles. In the future work, we will further introduce the corresponding work on the digital quadruplets in parallel driving. Also, the developing process of the realistic parallel driving system will be described in detail.

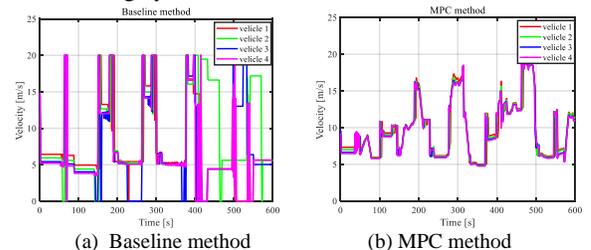
(a) Baseline method (b) MPC method
Fig. 11. MPC for fleet management in predictive vehicle.

## VII. CONCLUDING REMARKS

Under the enlightenment of ACP method, this paper presents the digital quadruplets in parallel driving in CPSS. These four components of digital quadruplets are called descriptive, predictive, prescriptive and real vehicles. The goal of the descriptive vehicles is modelling and describing the real vehicles and recording percepts from the real environment. The predictive vehicles aim to predict and plan for the automated in decision-making, controlling, planning and so on. The prescriptive vehicles are deciding how to interact with the real vehicles. Furthermore, the research process of the digital quadruplets in parallel driving is introduced. The presented digital quadruplets in parallel driving are expected to make the future connected automated driving safety, efficiently and synergistically. The framework of digital quadruplets for CPSS-based parallel driving still requires significant future research and development efforts along with their real-world verifications, which are being devoted by the research team.

## ACKNOWLEDGEMENT

This work was supported by Intel Collaborative Research Institute on Intelligent and Automated Connected Vehicles (ICRI-IACV), Program of Beijing Municipal Science & Technology Commission (Grant No. Z181100008918007), Key Program of National Natural Science Foundation of China (Grant No. 61533019) and NSFC National Key Program (Cognitive Computing of Visual and Auditory Information, Grant No. 91720000).